\documentclass{article}

\usepackage[preprint]{neurips_2024}

\usepackage[utf8]{inputenc} 
\usepackage[T1]{fontenc}    
\usepackage{hyperref}       
\usepackage{url}            
\usepackage{booktabs}       
\usepackage{amsfonts}       
\usepackage{nicefrac}       
\usepackage{microtype}      
\usepackage{xcolor}         
\usepackage{graphicx}
\usepackage{subcaption}
\usepackage{amsmath}

\title{Scaling Algorithm Distillation for Continuous Control with Mamba}

\author{
  Samuel Beaussant\\
  Akkodis Research\\
  France\\
  \texttt{samuel.beaussant@akkodis.com} \\
  \And
  Mehdi Mounsif\\
  Akkodis Research\\
  France\\
  \texttt{mehdi.mounsif@akkodis.com} \\
}

\begin{document}
\maketitle

\begin{abstract}
Algorithm Distillation (AD) was recently proposed as a new approach to perform In-Context Reinforcement Learning (ICRL) by modeling across-episodic training histories autoregressively with a causal transformer model. However, due to practical limitations induced by the attention mechanism, experiments were bottlenecked by the transformer's quadratic complexity and limited to simple discrete environments with short time horizons. In this work, we propose leveraging the recently proposed Selective Structured State Space Sequence (S6) models, which achieved state-of-the-art (SOTA) performance on long-range sequence modeling while scaling linearly in sequence length. Through four complex and continuous Meta Reinforcement Learning environments, we demonstrate the overall superiority of Mamba, a model built with S6 layers, over a transformer model for AD. Additionally, we show that scaling AD to very long contexts can improve ICRL performance and make it competitive even with a SOTA online meta RL baseline.
\end{abstract}


\section{Introduction}
Since its inception, the Transformer architecture \cite{Attention} has become the de facto model for many sequence modeling tasks across a vast range of domains. This widespread adoption can be partly explained by its impressive performance and versatility. However, its most striking feature is arguably the ability to perform "in-context" learning (ICL) after self-supervised pre-training on enough data. This departure from traditional in-weights learning has ignited significant interest as a novel form of meta-learning and made it a very good backbone for many foundation models \cite{DinoV2, SAM, vuong2023open}. In particular, Algorithm Distillation (AD) \cite{AD} leverages the in-context abilities of transformers to adapt and self-improve in new tasks from a few environment interactions. It has demonstrated very promising results from purely offline data in a meta reinforcement learning setting, but experiments have been limited to small and simple environments such as gridworlds. Hence, it is not clear if these appealing results could be further extended to more challenging and realistic environments with continuous state-action spaces.

One major difficulty arises from the fact that the context accessible to the meta-learner needs to be cross-episodic to enable long-range credit assignment and policy improvement. However, transformer models have been shown to perform poorly on benchmarks requiring modeling long-range dependencies \cite{LRA, LostInMiddle}. Moreover, by design, the attention mechanism powering its effectiveness scales quadratically with respect to the sequence length. As such, inference could be prohibitively slow for Reinforcement Learning (RL) tasks or even real-time Computer Vision tasks \cite{VisionMamba}.

Recently, State Space Models (SSMs) \citep{S4, Mamba} have outperformed transformer-based models in long-range sequence modeling tasks \citep{LRA}. In particular, the Mamba architecture based on the S6 layer \citep{Mamba} achieved impressive results in natural language processing (NLP) as well as DNA sequence and audio waveform modeling. Moreover, a recent work \citep{MambaICL} demonstrated that for simple supervised tasks, a Mamba model can exhibit similar ICL abilities as transformers. State Space Models also enjoy superior inference speeds and an efficient hardware implementation, making them potentially useful in many real-time control settings.

In this work, we study the synergy of Algorithm Distillation with Mamba for long-horizon tasks. We aim to investigate the impact of architecture inductive bias in this setting by comparing Mamba with a memory-efficient causal transformer \citep{FlashAttention}. Through four continuous control tasks, we show that, given similar model sizes, Mamba achieves better asymptotic performance in all settings considered. Furthermore, it seems that more complex tasks require longer contexts, which highlights the importance of efficiently modeling long sequences for optimal results when scaling AD. Finally, we compare the asymptotic performance of AD with two other SOTA meta-RL baselines and found that it is competitive with online meta-RL while training purely offline. It also outperforms prior SOTA offline meta-RL when compared in similar settings.

\section{Background}
\subsection{Meta Reinforcement Learning}
We consider the problem of \textbf{Meta Reinforcement Learning} and assume a distribution of tasks $p_{\mathcal{T}}$. Tasks are generally assumed to share a some fundamental similarities. Each task can be represented by a Markov Decision Process (MDP), defined as 5-tuple $\mathcal{T} = <S,A,P,R,\rho_0>$ where $S$ is the set of valid (continuous) states, $A$ is the set of valid (continuous) actions, $P(s_{t+1}|s_t, a_t)$ is the transition probability distribution describing the MDP dynamic, $R: S \times A \times S \mapsto \mathbb{R}$ is the reward function and $\rho_0$ is the initial state distribution. Usually in this framework, an agent (or decision-maker) is presented with a task ${\mathcal{T}}$ drawn from $p_{\mathcal{T}}$ during meta-training and must find a policy that maximizes its expected reward $G_t = \mathbb{E} \left[ \sum_{t}^T \gamma^t r_t\right]$. If meta-training succeeds, the meta-learner has learned the underlying structure and can find quickly a high-performing policy for a (potentially unseen) task $\mathcal{T}' \sim p_{\mathcal{T}}$. Ideally, adaptation to a new task (i.e meta-testing) should be possible from a small amount of data (referred to as few-shot learning). \textbf{In-Context Reinforcement Learning} is a family of methods dealing with Meta-RL. In this framework, adaptation to a new task happens entirely in the context of a sequence model, without any update to the model parameters. The term "in-context learning" was first coined in prior work on sequence models \citep{GPT3} to describe the meta-learning abilities acquired by large language models (LLMs) pre-trained on vast amount of data. It is opposed to "in-weights learning", i.e gradient-based learning with parameters updates. Depending on the sequence model used for adaptation, the task and the context length required, ICL performances may vary. 
\subsection{Mamba}
In this work, we leverage the Mamba architecture \citep{Mamba} based on the structured state-space models class \citep{S4}. This family of sequence models is related to RNNs, CNNs and classical state-space models. They map 1-D sequence $x_k \mapsto y_k \in \mathrm{R}$ using a latent state variable $h_k \in \mathrm{R}^N$ through the following relations:
\begin{equation}
    h_k = A h_{k-1} + B x_k
\end{equation}
\begin{equation}
    y_k = C h_k
\end{equation}
where $A \in \mathrm{R}^{N\times N}$, $B \in \mathrm{R}^{N\times1}$ and $C \in \mathrm{R}^{1\times N}$ are learnable parameters. When dealing with multi-dimensional data, each input channels is modeled by its own 1-D SSM. Importantly, the state matrix $A$ is structured i.e initialized to a diagonal matrix following the HiPPO theory \citep{Hippo} such that $A_{n,n} = -(n+1)$. Mamba is a fully recurrent model which enjoys fast inference, linear scaling in sequence length and an efficient hardware-aware implementation on GPU in order to be computationally efficient during training. Hence, it appears better suited than transformers in many reinforcement learning settings. 

\section{Algorithm distillation}
Broadly speaking, Algorithm Distillation tackles the meta-learning problem by learning to imitate a Reinforcement Learning algorithm. Over the course of training, an agent will learn to explore, plan, and correlate rewards to actions (also known as credit assignment). All of these complex behaviors are reflected in the learning data distribution of an agent. Modeling these histories using a sequence model thus entails learning to imitate these behaviors, which can later be repurposed during inference to in-context learn new tasks. This is similar to the few-shot learning abilities demonstrated by Large Language Models (LLMs) \citep{GPT3}. But instead of manually designed prompts or demonstrations, the model collects its own experience by interacting with the environment to bootstrap its own Reinforcement Learning in-context.  More formally, we call learning histories (or learning trajectories) the sequence of experiences collected by running a Reinforcement Learning algorithm $\mathcal{P}^{\textit{source}}$ on multiple individual tasks ${\mathcal{M}}^{N}_{n=1}$ producing dataset $D$:
\begin{equation}
    D := \{(o_0^{(n)}, a_0^{(n)}, r_0^{(n)}, ..., o_{T-1}^{(n)}, a_{T-1}^{(n)}, r_{T-1}^{(n)}, o_T^{(n)}, a_T^{(n)}, r_T^{(n)}) \sim \mathcal{P}^{\textit{source}}_{{\mathcal{M}}^{n}}\}^N_{n=1}
\end{equation}

Given a dataset $D$, a source algorithm can be distilled into a model by autoregressively predicting actions conditioned on a cross-episodic context. The original AD paper only considered discrete actions and thus trained their sequence model to map learning histories to probabilities over actions with a negative log likelihood (NLL) loss:
\begin{equation}
    L(\theta) := - \sum_{n=1}^N \sum_{t=1}^T \log P_\theta \left(A=a_t^{(n)} | h_{t-1}^{(n)}, o_t^{(n)}\right)
\end{equation}
with 
\begin{equation}
    h_t := (o_0, a_0, r_0, ..., o_{t-1}, a_{t-1}, r_{t-1}, o_t, a_t, r_t) = (o_{\leq t}, a_{\leq t}, r_{\leq t})
\end{equation}
a cross-episodic learning history. In this work, we consider only continuous-control problems. As such, we perform AD by optimizing the following loss:
\begin{equation}
    L(\theta) := \sum_{n=1}^N \sum_{t=1}^T  \left(a_t^{(n)} - f_\theta(h_{t-1}^{(n)}, o_t^{(n)})\right)^2
\end{equation}
The simple imitation-learning objective optimized during meta-training departs from traditional meta-RL methods that usually try to maximize the expected return $G_t$. The rationale justifying this behavior cloning loss is that correct action prediction given a learning history requires the model to approximate the policy improvement operator of the source algorithm. After the pre-training phase (i.e., the RL algorithm distillation), the sequence model can populate its own context by interacting with a new and unseen task and sequentially predicting actions. By imitating the source algorithm, it will effectively explore to gather diverse data at the beginning, assign credit to actions, and then slowly exploit successful policies.
\newline \newline
However, in practice, learning histories can be composed of millions of steps, which would require a significant amount of compute to model, particularly with causal transformers, which are quadratic in the sequence length. Rather than training on full sequences, we can keep one in every $k$ training episodes to effectively subsample the learning trajectories. The downsampling parameter $k$ is a hyper-parameter that needs to be tuned. Too high and the in-context learning becomes unstable as the learning histories are sparse. Too low significantly increases the compute burden and could even be detrimental to the sample efficiency of in-context learning. Additionally, one may also choose to randomly sample across-episodic subsequences $\bar{h_t} := (o_j, a_j, r_j, ..., o_{j+c}, a_{j+c}, r_{j+c})$ of length $c < T$ from $D$ rather than training with entire learning trajectories. However, as we show in section \ref{results}, this can lead to sub-optimal in-context learning or even prevent it.

\section{Related Works}
\label{sec:related_works}
In this section, we review works and methodologies broadly related to AD and Meta-RL. To better contextualize our own work, we divide the meta-RL domain into two categories, namely \textit{Context-based} Meta-RL and \textit{Gradient-based} Meta-RL. We also review some relevant work from the Offline RL field. 

\textbf{Context-based Meta Reinforcement Learning: }
Context-based (i.e., gradient-free) adaptation was first proposed and explored simultaneously in two concurrent works, $\text{RL}^2$ \citep{RL2} and \citep{wang2016learning}. Both approaches frame meta-learning as a Reinforcement Learning problem that can be solved using reinforcement learning. Using a memory-based neural network (such as an LSTM \citep{LSTM}), the meta-agent can discover the latent but shared structure underlying multiple related tasks to adapt quickly to unseen goals. Given the advantage of SSMs for sequence modeling compared to alternative models, a recent work \citep{metaS5} investigated the choice of S5 \citep{S5} as a drop-in replacement for online in-context meta-learning ($\text{RL}^2$) and found that it compares favorably to LSTMs while being faster than transformers. Additionally, the authors of \cite{PEARL} depart from the popular memory-based approach and propose using a probabilistic context variable encoded in a VAE's \cite{VAE} latent space to perform posterior sampling for adaptation. However, these works only consider the online Meta-RL setting, which is notoriously more expensive to run compared to the offline setting, as the meta-learner needs to interact with the environment at meta-training time. In our particular setting, AD requires a curated dataset of learning trajectories generated by source RL agents. However, meta-training (or pre-training) is still performed offline.

\textbf{Gradient-based Meta Reinforcement Learning: }
Gradient-based methods, unlike in-context approaches, achieve adaptation by fine-tuning the weights of the meta-learner at inference time (this is also sometimes referred to as \textit{in-weights} adaptation \citep{AD}). A number of works have explored this paradigm. Most notably, MAML \cite{MAML} learns a good parameter initialization point such that downstream adaptation requires as few gradient steps as possible. Many successor meta-learning works build on this framework. For instance, MACAW \citep{MACAW} adapts the popular offline RL algorithm Advantage-Weighted Regression \citep{AWR} to the offline Meta-RL setting by using MAML as their meta-optimizer. For online approaches, meta Q-Learning \citep{MQL} uses a simple context history encoded in a GRU \citep{GRU} along with an off-policy RL algorithm and was shown to outperform PEARL and ($\text{RL}^2$). Nevertheless, gradient-based and particularly MAML-flavored algorithms incur significant computational costs despite their effectiveness.

\textbf{Sequence Modeling for Offline Reinforcement Learning: }
Multiple works have recast Offline Reinforcement Learning as a sequence modeling problem with transformers. Most notably, Decision Transformer \citep{DT} demonstrated that transformers can learn high return policies from offline data. Concurrently, the Trajectory Transformer framework \citep{TT} was proposed to predict the dynamics of the environment for model-based planning. These Transformer-based approaches have demonstrated comparable or superior performance in standard Offline RL benchmarks. But none of these previous approaches demonstrated the emergence of in-context RL. Prompt Decision Transformer (PDT) \citep{PDT} also tackles the in-context offline Meta-RL setting. It investigates the ability of Transformer-based architecture to leverage medium to expert prompt demonstrations for few-shot learning. Their approach achieves new SOTA results in standard benchmarks. However, contrary to AD, Prompt DT requires at least medium-level demonstrations to adapt to a new task, defined as the middle 500 trajectories in the full replay buffer generated by the RL agents. If we consider a task that requires an agent to interact with a real physical system (say a robot), any extra learning on the physical system may requires human supervision, manual resetting and extensive care in how the robot operates to not damage itself or his surroundings. As such, reaching a "medium-level" performance in a test task with this setting to prompt a transformer could be prohibitively expensive or unpractical. To the best of our knowledge, only \citep{metaS5} considered SSMs for RL. This is likely due to the fact that current RL benchmarks are made up of environments with relatively short time horizons, hence not requiring long-range credit assignment. In contrast, AD natively requires modeling long-range dependencies and could benefit from SSMs.

\section{Experiments}
\label{sec:xp}
In this section, we investigate the in-context Reinforcement Learning performance of AD in multiple settings. When adapting to a test task, there is no fine-tuning nor gradient descent involved, contrary to the considered baselines. We pre-train our models with varying context lengths to study how it affects the performance of the downstream in-context learned policies. Following this, we perform a fair comparison between Mamba-based architectures and transformer-based models to assess the potential benefits of State-Space Models (SSMs). Both models have the same number of parameters and are trained using the same dataset. Finally, we explore AD as a competitive meta-RL approach and evaluate it against two prior SOTA methods, MQL \citep{MQL} and MACAW \citep{MACAW}. All experiments were run with a consumer-grade Nvidia RTX 3060 graphics card.

\subsection{Tasks and environments}
\label{sec:tasks}
We use slightly modified versions of four continuous-control RL benchmark environments, adapted to our meta-learning setting. We focus our study on robotic manipulation and locomotion. Just like the original AD paper, these environments support multiple tasks that cannot be solved in a zero-shot manner as observations are not sufficient to infer the task. These prerequisites ensure that the agent has to adapt to an unseen task by trial-and-error. However, unlike previous work, we consider tasks with a longer time horizon which makes across-episodic history modeling more challenging. This addresses one of the limitations acknowledged by the authors of AD. The four environments are briefly described below:

\textbf{Reacher-Goal:} A 2-link robot arm must reach a 2D target in an unknown location inside a circle. As such, the agent needs to explore and perform credit assignment to successfully solve the task. Each episode is 50 timesteps long.

\textbf{Pusher-Goal:} A 7 DoF human-like robot arm has to push a target cylinder to a target location using its end-effector. Both the target cylinder location and the target location are unknown and must be in-context learned. The episodes last 100 timesteps.

\textbf{Half-Cheetah-Vel:} A multi-jointed 2D cheetah-like robot must run at a desired but not observed speed. Once again, exploration and credit assignment are mandatory to learn a successful policy. Episodes are 200 steps long.

\textbf{Ant-Dir:} A high-dimensional locomotion task where a 3D robot has to coordinate its four legs to run as fast as possible in the goal directions (randomly sampled between 0 and 360 degrees). The goal is concealed from the agent and must be inferred through interactions with the environment. Episodes are 200 steps long.

Both Half-Cheetah-Vel and Ant-Dir have already been explored in previous works such as \citep{MAML, MACAW, MQL, PEARL} and are popular benchmark tasks for meta-RL. In this work, we introduce the Pusher-Goal environment to provide a hard exploration problem and challenge the in-context learning abilities of AD. All tasks are run using the Mujoco simulator \citep{Mujoco} with OpenAI Gym \citep{gym}. 

\subsection{Pre-training}
Algorithm Distillation is a two-step process involving first data generation followed by autoregressive model training.

\textbf{Data generation:}
For each environment, a learning trajectory is generated by running a SOTA Reinforcement Learning algorithm on a uniformly sampled task. More specifically, we used Proximal Policy Optimization (PPO) \cite{PPO}, Soft Actor Critic \citep{SAC}, and DroQ \citep{DroQ} depending on which was more appropriate for a particular MDP, both in terms of sample efficiency and training stability. For instance, PPO was very stable (i.e increases the mean return monotonically) but converges slower than off-policy algorithms such as SAC and DroQ. As such the length of the learning trajectories also increases which incur additional cost during pre-training. In contrast, both SAC and DroQ are very sample-efficient but unstable in some cases, with random and sharp performance volatility. This unstable behavior might also emerged during ICRL (as AD imitate the RL algorithm) and decrease performance. Hence the choice to chose different algorithms depending on the environment to find a trade-off between stability and sample-efficiency.  Characterizing the impact of data distribution is an interesting and important question which let for future work. Exact hyper-parameters, including the downsampling parameter $k$ and dataset-related technical details, can be found in Appendix \ref{appendix_data}. After data collection, the learning trajectories of the collected tasks are downsampled using the same environment-specific factor $k$. At training time, we sample a batch containing multiple (potentially sub)sequences of different tasks and autoregressively predict the continuous actions. 

\textbf{Model training:}
We compare Mamba with a similarly sized Decision Transformer trained on the same data and with the same context length. Exact parameter counts, hyper-parameters, and architecture can be found in Appendix \ref{appendix_models}. Both models output continuous actions conditioned on a sequence of tokens. Unlike the original AD work, we do not consider that each token represents either a state, an action, or a reward. Instead, we concatenate one-step transitions into a single token such that $c^i = (s_{t}^i, a_{t}^i, r_{t}^i, s_{t+1}^i)$. In other words, transition tuples are seen as a single token. In our preliminary experiments, both approaches were competitive, but the former entails using a context $3 \times$ longer, which exceeds our compute budget for all but the Reacher-Goal environment. To improve robustness, we add a small amount of isotropic Gaussian noise to each token of the input sequences.

\subsection{Results}
\label{results}
We analyze AD's adaptation efficacy and focus our discussion on the sample-efficiency and asymptotic performance (i.e., mean return) metrics. We also report the asymptotic performance of both baselines on the same benchmark. For all of our evaluations, we aggregate the performance over 3 pre-training random seeds and for 10 unseen test tasks each.
\begin{figure*}[!ht]
     \centering
     \begin{subfigure}{0.48\textwidth}
         \centering
         \includegraphics[width=1\textwidth]{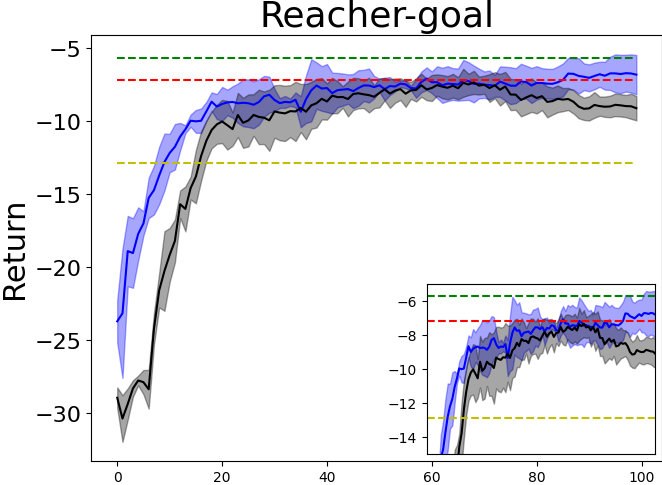}
         \label{reacher}
     \end{subfigure}
     \hspace{0.3cm}
     \begin{subfigure}{0.48\textwidth}
         \centering
         \includegraphics[width=1\textwidth]{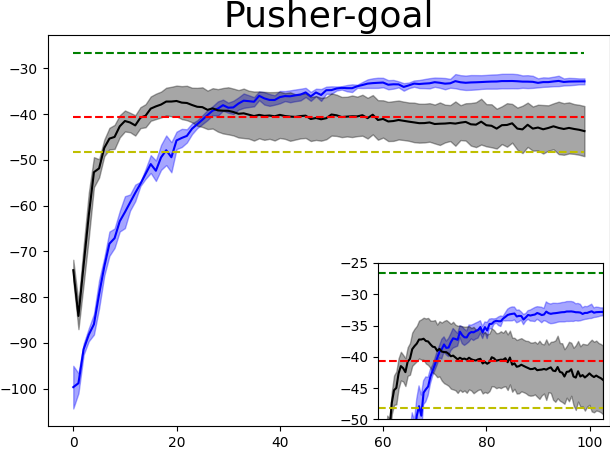}
         \label{pusher}
     \end{subfigure}
     \par\bigskip
     \begin{subfigure}{0.48\textwidth}
         \centering
         \includegraphics[width=1\textwidth]{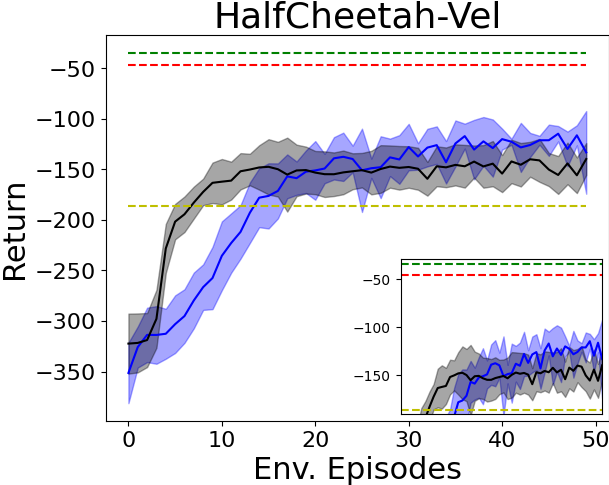}
     \end{subfigure}
     \hspace{0.3cm}
     \begin{subfigure}{0.48\textwidth}
         \centering
         \includegraphics[width=1\textwidth]{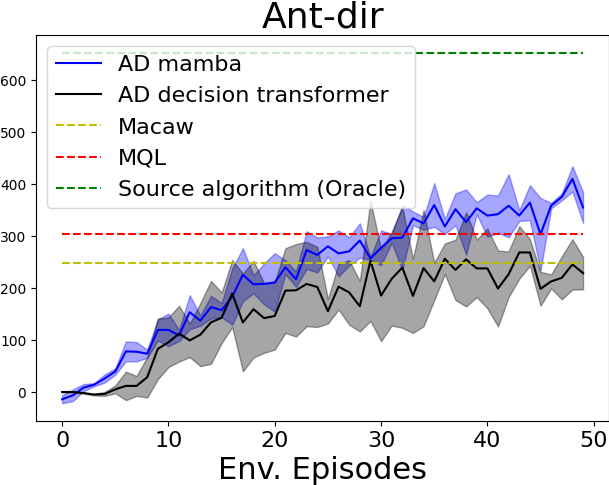}
     \end{subfigure}
        \caption{Main results: each learning curve is obtained by pre-training and testing the models with the best performing context length on validation tasks. We evaluate AD with Mamba and a Decision Transformer architecture against MACAW and MQL asymptotic performance.}
        \label{inference_xp_perf}
\end{figure*}
\newline\newline
\textbf{What is the best suited architecture for AD ?}  Figure~\ref{inference_xp_perf} reports the in-context RL performance for both a Mamba-based architecture and a similarly sized Decision Transformer. More specifically, we plot the best performing context length for each model. As shown, the Mamba variant outperforms the Transformer variant on all the environments. Differences in performance range from a small edge in Reacher-Goal and Half-Cheetah-Vel to a more significant improvement in Ant-Dir and Pusher-Goal. Interestingly, using a Decision Transformer with AD leads to the worst results amongst all baselines in Ant-dir, which is arguably the most complex environment. We believe that this could be evidence for Transformer-based AD not scaling well above a certain task complexity. These results are consistent with multiple other works that found that SSMs are better suited for modeling long-range dependencies, as required for credit assignment. Moreover, from a practical standpoint, we found Mamba to be easier to train, requiring less hyper-parameter tuning and being overall more stable during training. 

\begin{figure*}[!ht]
     \centering
     \begin{subfigure}{0.48\textwidth}
         \centering
         \includegraphics[width=1\textwidth]{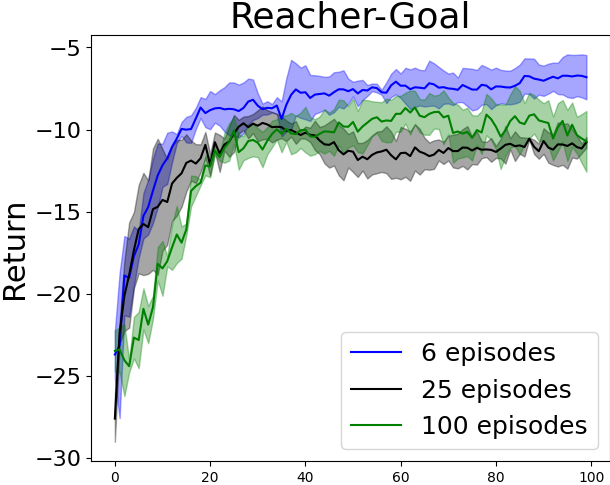}
     \end{subfigure}
     \hspace{0.3cm}
     \begin{subfigure}{0.48\textwidth}
         \centering
         \includegraphics[width=1\textwidth]{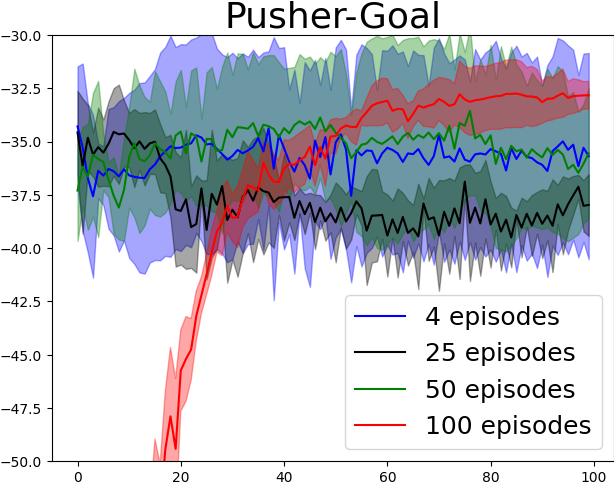}
     \end{subfigure}
     \par\bigskip
     \begin{subfigure}{0.48\textwidth}
         \centering
         \includegraphics[width=1\textwidth]{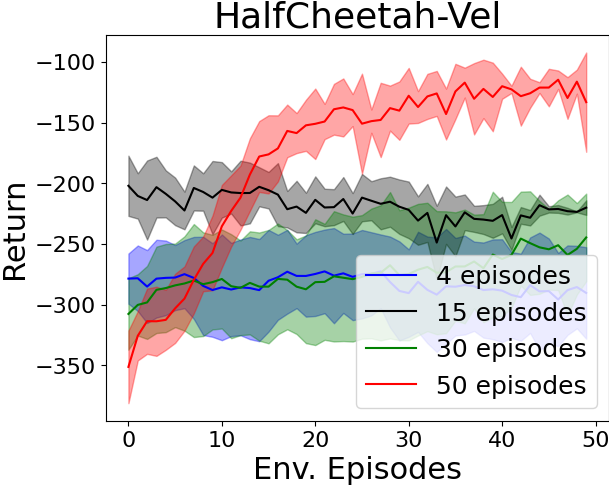}
     \end{subfigure}
     \hspace{0.3cm}
     \begin{subfigure}{0.48\textwidth}
         \centering
         \includegraphics[width=0.99\textwidth]{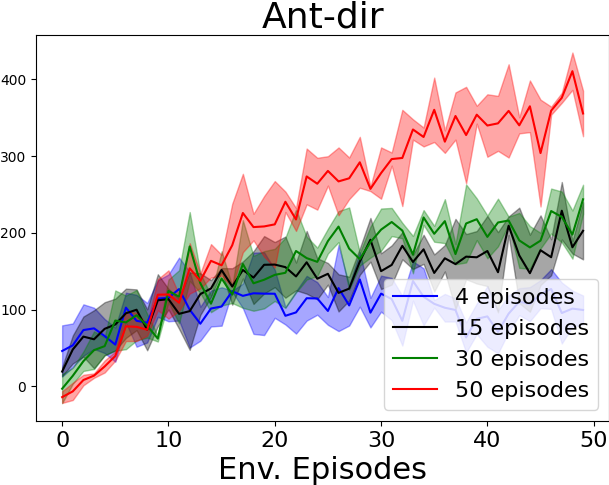}
     \end{subfigure}
        \caption{Impact of context size on ICRL performance. Each learning curve is obtained by pre-training and testing a Mamba model with a particular context length on validation tasks. Except for the simpler Reacher-goal environment, context size seems correlated with task complexity.}
        \label{inference_xp_ctx_length}
\end{figure*}

\textbf{How does context length impact ICL performance ?} In-context RL was shown to emerge with AD when the context size was large enough, spanning across episodes, to effectively capture improvements in the training data. Prior work concluded that a 2-4 episodes long (200 steps) context was enough to learn a near-optimal in-context RL algorithm but did not study longer contexts. Besides, the experiments were conducted on relatively simple discrete MDPs. Therefore, we wonder how these observations may evolve when dealing with more challenging environments. We re-investigate this matter by training several AD variants with different context lengths. To go a step further, we study how very long context sizes, ranging from a few episodes to entire (sub-sampled) learning trajectories, affect downstream in-context RL. Results for the Mamba-based architecture are plotted in Figure~\ref{inference_xp_ctx_length}.

Nearly all the context sizes experimented with enable in-context RL, except for the Half-Cheetah-Vel task where only AD with a context containing the (down-sampled) full history seems to effectively learn. For all other smaller context sizes, the AD variants underperform and seem to plateau after trying to zero-shot the task. Even then, performance correlates positively with context length. Although the best performing AD variants in Ant-Dir have the longest context, shorter context lengths also exhibit in-context RL but to a lesser extent. Surprisingly, in Pusher-Goal, shorter context sizes do very well in terms of sample efficiency and asymptotic performance, nearly zero-shotting the new tasks. Still, AD with full context performs better as it seems to explore more extensively. On the other hand, for the simpler Reacher-Goal environment, a shorter context size results in more efficient in-context RL. We hypothesize that due to the relative simplicity of the task, long-range credit assignment may not be required. Overall, the optimal context length seems related to the task complexity, which could be defined in terms of the state-action space dimensionality and the task time-horizon.

\textbf{How does AD compares to other Meta-RL approaches ?} We perform a comparative study of AD adaptation efficiency against MACAW and MQL, two off-policy SOTA meta-RL methods. Rigorously, both are not directly comparable to AD due to incompatible algorithmic designs. For instance, MQL learns online and gets to interact with the environment at meta-training time, unlike MACAW and AD, which both learn from offline data. Nonetheless, our aim is instead to roughly situate the AD approach in relation to other SOTA methods when it comes to overall adaptation efficiency. As such, we analyze their respective asymptotic performance in the most similar configurations possible. As such, MACAW is trained on data coming from the same dataset as AD, and MQL is trained online on the same number of tasks. Results are reported in Figure~\ref{inference_xp_perf}. For AD, we plot the variants with the best performing context length. As a reference, we also plot on the same figure the asymptotic performance of the source RL algorithm. This baseline provides an 'oracle' for asymptotic performance. Both MQL and AD are close to optimal performance in Reacher-Goal. But the gap widens in all other environments except for MQL in Half-Cheetah-Vel. Surprisingly, AD is competitive with MQL even though it was trained fully offline. It even outperforms it on 2 out of 4 tasks but struggles on Half-Cheetah-Vel. Regarding MACAW, AD systematically achieves higher mean returns on all environments. It is worth mentioning that MACAW was trained with significantly less data than in its original paper, which explains its relative under-performance. Indeed, in \cite{MACAW}, MACAW was meta-trained with 45 tasks and evaluated on 5. We found this experimental setting to be very unbalanced and not challenging enough to reflect the extrapolation and adaptation capabilities of AD and other approaches. As such, we limited the amount of meta-training data. For instance, we used 30 tasks during training for our Ant-dir experiments and tested on 30 unseen direction. Similarly, we used only 15 velocities with HalfCheetah-vel. 
These results highlight that, while being very simple, AD is a competitive Meta-RL approach when used in conjunction with a powerful sequence model such as Mamba.

\section{Conclusion}
\label{sec:conclusion}
In this work, we addressed arguably one of the main limitations of AD as proposed in the original paper and extended the range of possible applications. We showed that Mamba, a recently proposed SSM, can efficiently handle multi-episodic context and exhibit ICRL when powering AD in complex environments. Experimentally, AD Mamba demonstrated clear advantages over our baseline Decision Transformers, achieving better performance while also being more scalable. Furthermore, experiments showcased that modeling longer contexts tends to improve ICRL on more complex tasks, highlighting the benefit that powerful long-horizon sequential models can bring to AD. This relates to recent findings on many-shot ICL for NLP tasks \cite{ManyShots}. Finally, we showed that the synergy between Mamba and AD enables it to outperform previous offline SOTA approaches and to compete with MQL.

\textbf{Limitations and Future Work:} Despite appealing results from only offline data, some limitations remain. This study considered widely used Meta-RL environments to benchmark AD. However, these environments are still manageable for a transformer-based AD (even if it barely learn in Ant-dir). Evaluating AD on environments with potentially very long horizon and/or which requires extensive exploration is worth while and an interesting extension to this work. Gathering enough learning trajectories to distill a reinforcement learning algorithm could be impractical. Even though the initial cost is amortized by producing a sample-efficient in-context RL algorithm, this procedure may not be suited for applications where online learning is prohibitively expensive. Future work could investigate how AD may leverage existing demonstration datasets that perhaps do not contain enough behavioral diversity (e.g., exploration-exploitation trade-offs) for ICRL. Additionally, future work may explore how the data and the sampling scheme used for pre-training, shape the ICLR abilities. Indeed, a contemporary study \cite{SkillIt} showed that skill or behavior acquisition during pre-training can be impacted by how the data is sampled. Similarly, it could be interesting to research how to mix and sample beginner/medium/expert trajectories for optimal ICRL with a data-driven approach. On the same note, incorporating prompting strategies during inference to improve the sample efficiency of ICRL is also an exciting direction of research. Finally, recent work \citep{RL_diffusion} unveiled the strong performance of diffusion models for behavioral cloning. As such, replacing the simple autoregressive-based distillation with a diffusion-based approach may improve the performance and generalization ability of AD.
\newpage

\bibliography{neurips_2024.bib}  


\appendix
\newpage
\section{Dataset Generation}
\label{appendix_data}

\begin{table}[!ht]
  \centering
  \begin{minipage}[t]{0.32\textwidth}
    \caption{PPO hyper-parameters}
    \label{table:ppo}
    \centering
    \begin{tabular}{ll}
      \toprule
      Parameter     & Value \\
      \midrule
      num. epochs & 10  \\
      steps per env.     & 500      \\
      num. envs     & 5  \\
      entropy coeff. & 5e-3 \\
      learning rate & 1e-4 \\
      batch size & 64 \\
      gamma & 0.99 \\
      gae lambda & 0.95 \\
      epsilon (clip) & 0.2 \\
      value coeff & 0.5 \\
      \bottomrule
    \end{tabular}
  \end{minipage}%
  \hfill
  \begin{minipage}[t]{0.32\textwidth}
    \caption{SAC hyper-parameters}
    \label{table:sac}
    \centering
    \begin{tabular}{ll}
      \toprule
      Parameter     & Value \\
      \midrule
      buffer size & 1e6  \\
      num warm-up steps     & 400      \\
      num. envs     & 5  \\
      learning rate & 1e-4 \\
      batch size & 256 \\
      gamma & 0.99 \\
      gae lambda & 0.95 \\
      epsilon (clip) & 0.2 \\
      value coeff & 0.5 \\
      \bottomrule
    \end{tabular}
  \end{minipage}%
  \hfill
  \begin{minipage}[t]{0.32\textwidth}
    \caption{DroQ hyper-parameters}
    \label{table:droq}
    \centering
    \begin{tabular}{ll}
      \toprule
      Parameter     & Value \\
      \midrule
      buffer size & 1e6  \\
      num. warm-up steps     & 400      \\
      num. gradient steps & 3 \\
      policy update freq. & 3 \\
      num. envs     & 5  \\
      learning rate & 1e-4 \\
      batch size & 256 \\
      gamma & 0.99 \\
      gae lambda & 0.95 \\
      epsilon (clip) & 0.2 \\
      value coeff & 0.5 \\
      \bottomrule
    \end{tabular}
  \end{minipage}
\end{table}

\begin{table}[!ht]
  \caption{Environments hyper-parameters}
  \centering
  \begin{tabular}{lcccc}
    \toprule
    Settings & Reacher-goal & Pusher-goal & HalfCheetah-vel & Ant-goal \\
    \midrule
    RL algorithm & SAC  & PPO & SAC & DroQ  \\
    num. train tasks & 70 & 35 & 15 & 30 \\
    Train steps per task & 1e5 & 5e5 & 4e5 & 5e5 \\
    \bottomrule
  \end{tabular}
\end{table}

\newpage
\section{Models pre-training}
\label{appendix_models}

\begin{table}[!ht]
  \caption{Common hyper-parameters}
  \centering
  \begin{tabular}{lcccc}
    \toprule
    Settings & Reacher-goal & Pusher-goal & HalfCheetah-vel & Ant-goal \\
    \midrule
    Optimizer & \multicolumn{4}{c}{Adam}  \\
    Epochs & \multicolumn{4}{c}{450}  \\
    Learning rate Scheduler & \multicolumn{4}{c}{cosine with warmup} \\
    Downsampling rate k & 4  & 10 & 8 & 10 \\
    weight decay & 1e-4 & 2e-4 & 5e-4 & 5e-4\\
    \bottomrule
  \end{tabular}
\end{table}

\begin{table}[!ht]
  \caption{Mamba hyper-parameters}
  \centering
  \begin{tabular}{lcccc}
    \toprule
    Settings & Reacher-goal & Pusher-goal & HalfCheetah-vel & Ant-goal \\
    \midrule
    Embedding dim. & \multicolumn{4}{c}{32}  \\
    Dim. model & 384 &  512 & 512 & 512 \\
    num. layers & 6 & 8 & 8 & 8\\
    num. params. & 6M & 12M & 12M & 12M \\
    \bottomrule
  \end{tabular}
\end{table}

\begin{table}[!ht]
  \caption{Decision Transformer hyper-parameters}
  \centering
  \begin{tabular}{lcccc}
    \toprule
    Settings & Reacher-goal & Pusher-goal & HalfCheetah-vel & Ant-goal \\
    \midrule
    num. heads & \multicolumn{4}{c}{4} \\
    pos. encoding & \multicolumn{4}{c}{Absolute} \\
    feed fwd. dim. & \multicolumn{4}{c}{2048} \\
    Embedding dim. & 42 & 64 & 64 & 128 \\
    num. layers & 8 & 9 & 9 & 6\\
    num. params. & 6M & 12M & 12M & 14M \\
    \bottomrule
  \end{tabular}
\end{table}

\end{document}